%% file: main.tex
\documentclass[runningheads]{llncs}
\usepackage[utf8]{inputenc}

\usepackage{eccv}

\usepackage{eccvabbrv}

\usepackage{graphicx}
\usepackage{booktabs}
\usepackage{tikz}
\usetikzlibrary{arrows.meta,calc,positioning,fit,backgrounds}
\usepackage{graphicx}
\usepackage{multirow}

\usepackage[accsupp]{axessibility}  

\usepackage{hyperref}

\usepackage{orcidlink}

\begin{document}

\title{EgoGazeLite: On-Device Egocentric Gaze Prediction for Token-Efficient Multimodal LLM Video Input} 

\titlerunning{EgoGazeLite: Egocentric Gaze Prediction for MLLMs}

\author{Matteo Stoiber\orcidlink{0009-0005-8689-3341}\inst{1} \and
Niels Buus Lassen\inst{1}}

\authorrunning{M.~Stoiber and N.~B.~Lassen}

\institute{Copenhagen Business School, Department of Digitalization, Copenhagen, Denmark\\
\email{matteostoiber@gmail.com}, \email{nbl.digi@cbs.dk}}

\maketitle

\begin{abstract}
 The use of multimodal LLMs (MLLMs) for egocentric video understanding with wearable devices is constrained by the token budget. Memory and compute cost scale with the number of visual tokens, and high-resolution video quickly becomes expensive to transmit and process at scale. Prior work (GazeLLM) addresses this by cropping the video around the camera wearer's gaze. This reduces the number of visual tokens by about tenfold while maintaining or improving the quality of full-resolution descriptions. However, this compression strategy depends on dedicated eye-tracking hardware, which is unavailable on consumer smart glasses. Building a software-only substitute poses a joint constraint: the predictor must be accurate enough to preserve downstream description quality, yet light enough to run on-device, within the power and compute budget of a smartphone. We address this with EgoGazeLite, a lightweight dual-process gaze predictor for egocentric video. Across two MLLMs, three automated metrics, and two LLM judges, predicted-gaze crops show no significant difference from ground-truth-gaze crops. Equivalence is confirmed in all ten cases. EgoGazeLite achieves this at 15.7M parameters, 6.71 GFLOPs, and runs the full gaze-and-crop pipeline end-to-end in real time (21.6\,ms/frame) on consumer accelerator hardware. Together, these results remove the need for eye-tracking hardware for token-efficient, gaze-conditioned egocentric video understanding with MLLMs.
 
\end{abstract}

\input{figures/fig1_teaser.tex}

\input{sections/01_intro}
\input{sections/02_related}
\input{sections/03_method}
\input{sections/04_experiments}
\input{sections/05_conclusion}

\section*{Acknowledgements}
All computation for this project was performed on UCloud, the Danish national interactive HPC system operated by the Interactive HPC Consortium (University of Southern Denmark, Aarhus University, and Aalborg University). The resources used were allocated by Copenhagen Business School.

%
%
\bibliographystyle{splncs04}
\bibliography{main}
\end{document}

%% file: figures/fig1_teaser.tex
\definecolor{colorPredicted}{HTML}{FF7F0E}
\definecolor{colorGT}{HTML}{1F77B4}
\definecolor{colorCenter}{HTML}{7F7F7F}
\definecolor{colorBackbone}{HTML}{9467BD}
\definecolor{colorSaliency}{HTML}{1F77B4}
\definecolor{colorAttention}{HTML}{00A8CC}
\definecolor{colorFusion}{HTML}{FF7F0E}
\definecolor{colorError}{HTML}{D62728}
\definecolor{colorMLLM}{HTML}{1A3A6B}

\tikzset{
    egoBlock/.style={
        draw=colorBackbone,
        fill=colorBackbone!10,
        thick,
        rounded corners=3pt,
        inner sep=5pt,
        font=\scriptsize\sffamily\bfseries
    }
}

\newcommand{\imgPrev}{eccv_assets/fig1_frame_prev.jpg}
\newcommand{\imgCurrClean}{eccv_assets/fig1_frame_curr_clean.jpg}
\newcommand{\imgCurrOverlay}{eccv_assets/fig1_frame_curr_overlay.jpg}
\newcommand{\imgCrop}{eccv_assets/fig1_crop_predicted.jpg}

\newcommand{\stagelabel}[2]{%
    \node[font=\tiny\sffamily, text=colorCenter,
          anchor=north, yshift=-4pt] at (#1) {#2};}

\newcommand{\tritipX}[4]{%
    \coordinate (tritip) at ([xshift=#3] #2);
    \fill[#1]
        (tritip) --
        ([xshift=-0.15cm, yshift= #4] tritip) --
        ([xshift=-0.15cm, yshift=-#4] tritip) -- cycle;}

\begin{figure}[h!]
\centering
\begin{tikzpicture}[>=Stealth, thick]


\def\thumbW{1.0cm}      
\def\stackOff{0.13cm}   
\def\mainW{2.5cm}       
\def\cropW{1.4cm}       
\def\encLen{0.80cm}     
\def\encTall{0.72cm}    
\def\encShort{0.48cm}   
\def\mllmW{1.05cm}      
\def\mllmH{1.50cm}      

\def\colAB{0.35cm}      
\def\colBC{0.35cm}      
\def\colCD{0.35cm}      
\def\colEncTok{0.35cm}  
\def\colTokW{0.22cm}    
\def\colTokMllm{0.35cm} 

\def\triAB{0.22cm}      
\def\triBC{0.30cm}      
\def\triCD{0.25cm}      
\def\triEncTok{0.22cm}  
\def\triTokMllm{0.25cm} 

\def\triSmall{0.09cm}
\def\triLarge{0.12cm}

\node[anchor=south west] (framePrev) at (\stackOff,\stackOff) {%
    \includegraphics[width=\thumbW,height=\thumbW,
                     keepaspectratio=false]{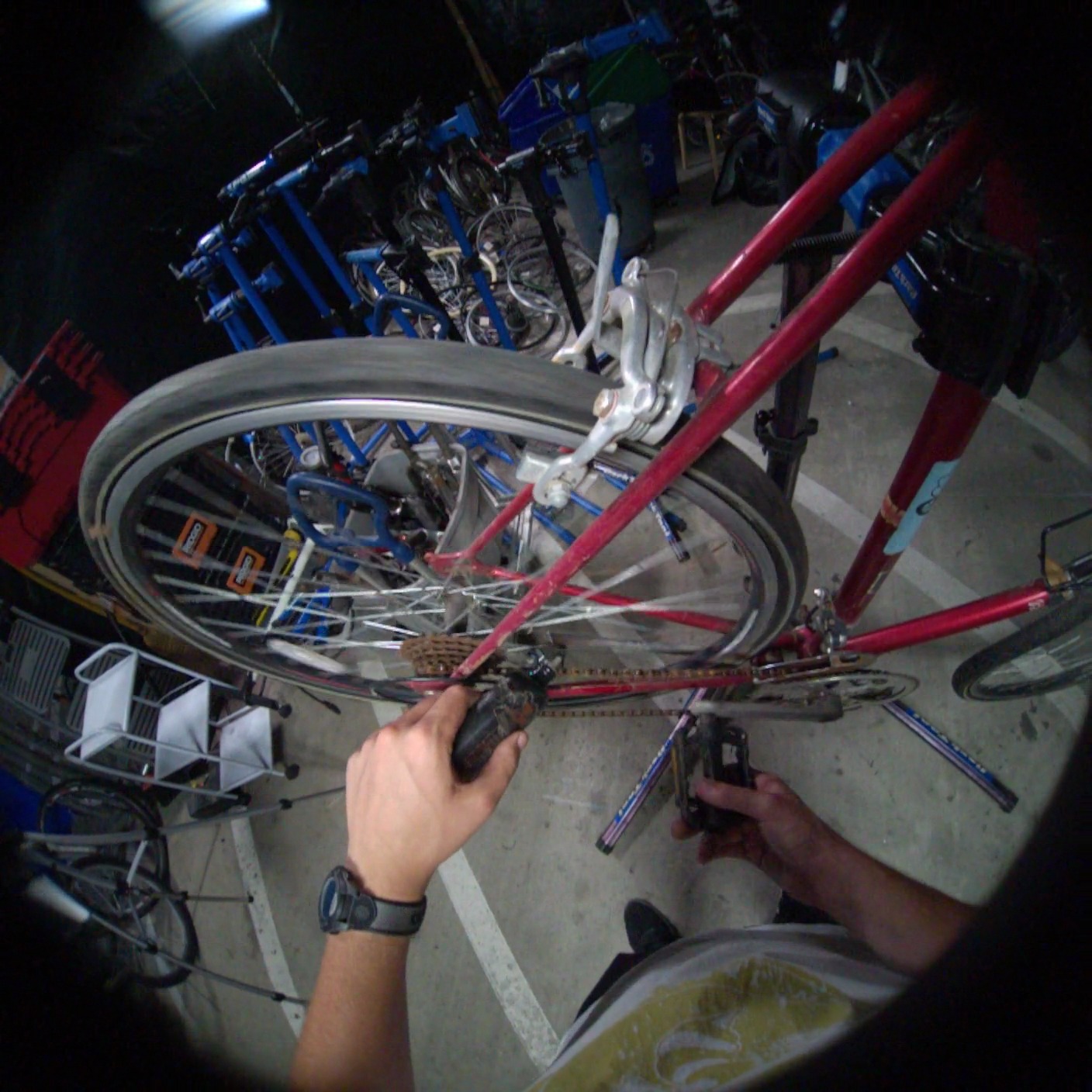}};

\node[anchor=south west] (frameCurr) at (0,0) {%
    \includegraphics[width=\thumbW,height=\thumbW,
                     keepaspectratio=false]{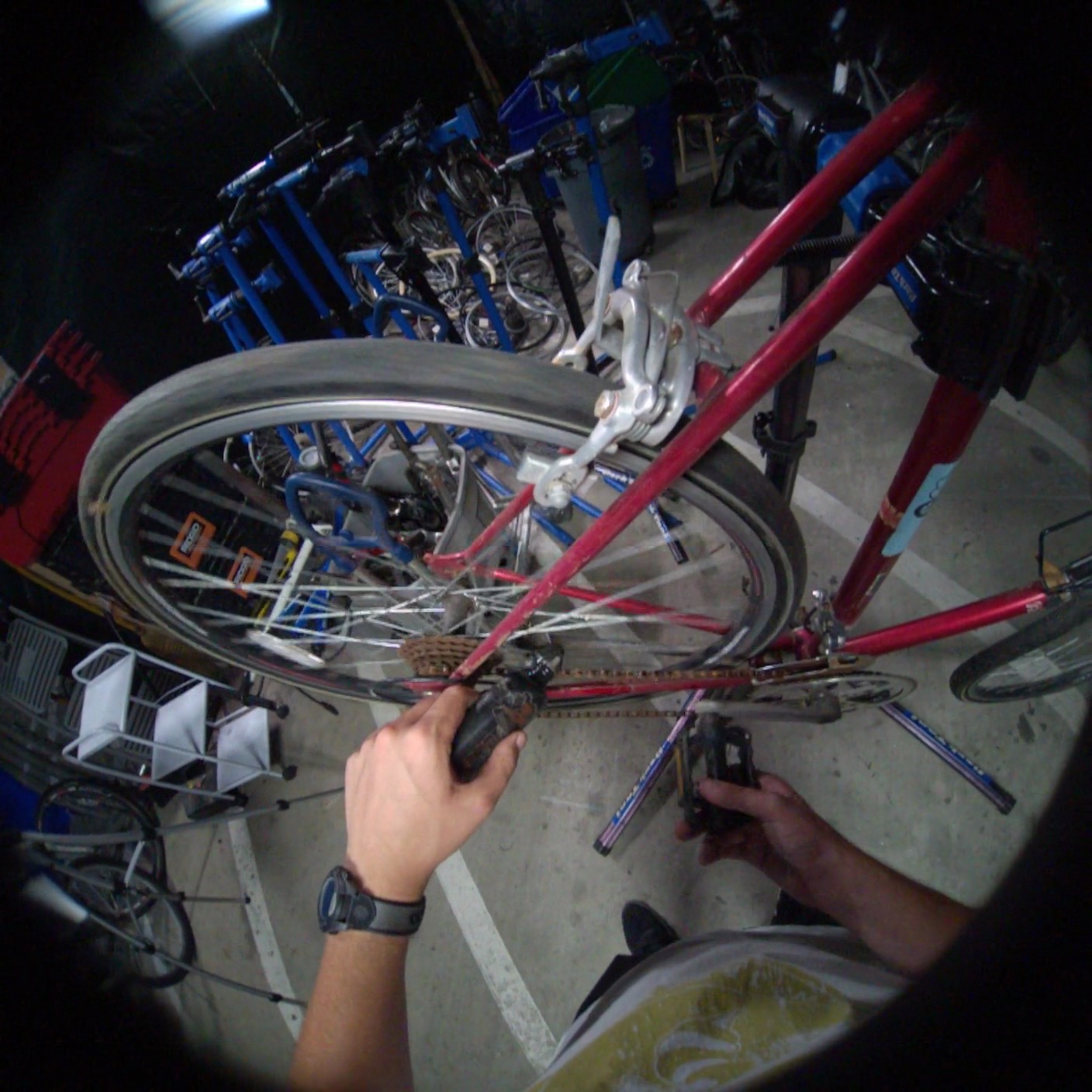}};

\stagelabel{frameCurr.south}{Egocentric video}

\coordinate (z1out) at
    ($(frameCurr.east)!0.5!(framePrev.east)+(0,0.5*\stackOff)$);

\tritipX{colorFusion}{z1out}{\triAB}{\triSmall}

\node[egoBlock,
      anchor=west,
      minimum width=1.45cm,
      minimum height=1.1cm,
      align=center,
      xshift=\colAB] (egoBox) at (z1out) {%
    EgoGazeLite};

\stagelabel{egoBox.south}{Gaze prediction}

\tritipX{colorCenter}{egoBox.east}{\triBC}{\triLarge}

\node[anchor=west, xshift=\colBC] (mainFrame) at (egoBox.east) {%
    \includegraphics[width=\mainW,height=\mainW,
                     keepaspectratio=false]{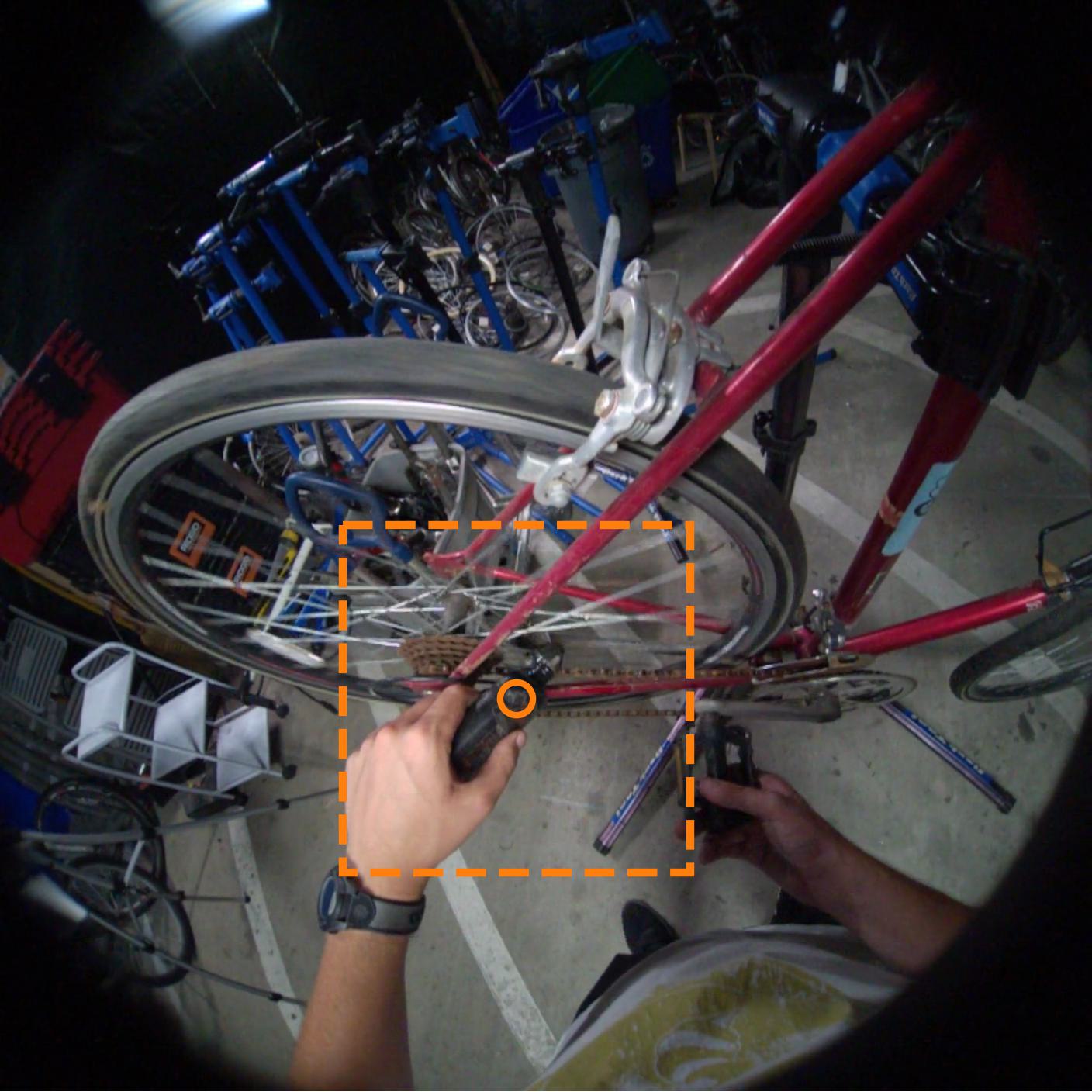}};

\node[font=\tiny\sffamily\bfseries, text=white,
      anchor=south west, xshift=3pt, yshift=3pt]
    at (mainFrame.south west) {1\textsuperscript{st} person view};

\node[anchor=west, xshift=\colCD] (cropThumb) at (mainFrame.east) {%
    \includegraphics[width=\cropW,height=\cropW,
                     keepaspectratio=false]{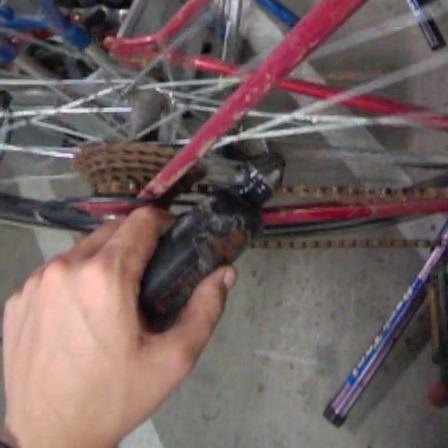}};

\draw[colorPredicted, line width=1.4pt]
    (cropThumb.north west) rectangle (cropThumb.south east);

\stagelabel{cropThumb.south}{Gaze-focused video}

\def\zoomTopFrac{0.68}
\def\zoomBotFrac{0.22}
\coordinate (zoomT) at
    ($(mainFrame.south east)!{\zoomTopFrac}!(mainFrame.north east)$);
\coordinate (zoomB) at
    ($(mainFrame.south east)!{\zoomBotFrac}!(mainFrame.north east)$);
\draw[colorPredicted, thin] (zoomT) -- (cropThumb.north west);
\draw[colorPredicted, thin] (zoomB) -- (cropThumb.south west);

\tritipX{colorFusion}{cropThumb.east}{\triCD}{\triSmall}

\coordinate (encL) at ([xshift=\colCD] cropThumb.east);

\coordinate (encTL) at ($(encL)+(0,\encTall)$);
\coordinate (encBL) at ($(encL)+(0,-\encTall)$);
\coordinate (encTR) at ($(encL)+(\encLen,\encShort)$);
\coordinate (encBR) at ($(encL)+(\encLen,-\encShort)$);

\fill[colorSaliency!20] (encTL)--(encTR)--(encBR)--(encBL)--cycle;
\draw[colorSaliency, thick] (encTL)--(encTR)--(encBR)--(encBL)--cycle;

\node[font=\tiny\sffamily, align=center]
    at ($(encL)+(0.45*\encLen,0)$) {Video\\encoder};

\coordinate (encMid) at ($(encTR)!0.5!(encBR)$);

\tritipX{colorCenter}{encMid}{\triEncTok}{\triSmall}

\coordinate (tokBL) at ([xshift=\colEncTok, yshift=-0.70cm] encMid);
\coordinate (tokTR) at
    ([xshift=\colEncTok, yshift=-0.70cm] encMid);
\coordinate (tokTR) at
    ($(tokBL)+(\colTokW, 1.40cm)$);

\foreach \i in {0,1,2,3,4,5,6,7}{
    \pgfmathsetmacro{\yb}{0.02 + \i * 0.168}
    \fill[colorFusion!80]
        ($(tokBL)+(0.04cm,\yb cm)$) rectangle +(0.14cm,0.14cm);
}
\draw[colorCenter, thick, dashed, rounded corners=1pt]
    (tokBL) rectangle (tokTR);

\coordinate (tokR) at ($(tokTR)!0.5!(tokTR |- tokBL) + (0, 0.70cm)$);
\coordinate (tokR) at ([yshift=0cm] tokTR |- encMid);

\tritipX{colorCenter}{tokR}{\triTokMllm}{\triSmall}

\node[anchor=west,
      draw=colorMLLM,
      fill=colorMLLM,
      thick,
      rounded corners=4pt,
      minimum width=\mllmW,
      minimum height=\mllmH,
      align=center,
      font=\scriptsize\sffamily\bfseries,
      text=white,
      xshift=\colTokMllm] (mllmBox) at (tokR) {Multi\\modal\\LLM};

\end{tikzpicture}
\caption{EgoGazeLite removes GazeLLM's eye-tracking dependency, enabling gaze-guided video cropping on edge devices while reducing MLLM input by 90\%.}
\label{fig:teaser}
\end{figure}
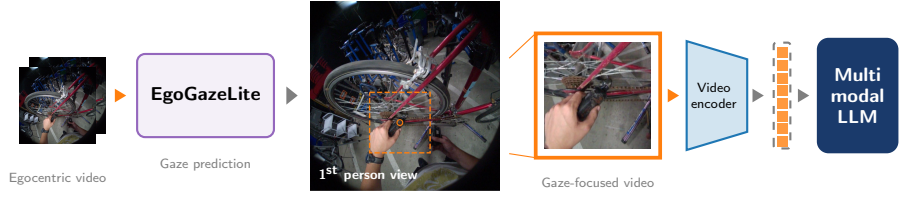

%% file: sections/01_intro.tex
\section{Introduction}
\label{sec:intro}

Multimodal large language models (MLLMs) are increasingly capable of understanding first-person, egocentric videos. This opens the door to systems that can observe a person performing a complex task, such as cooking, surgery, or equipment repair, and generate accurate descriptions or answer questions about it. These systems could have significant implications for skill transfer, real-world task guidance, and assistive technology for individuals with visual impairments.

However, realizing this promise at scale runs into a hard token-budget constraint. MLLMs typically encode visual input through Vision Transformers~\cite{dosovitskiy2020vit}, whose memory cost scales linearly with the number of visual tokens, which in turn scales with resolution and duration. Reducing this token budget without losing task-relevant detail is the central efficiency challenge. 

Rekimoto's GazeLLM~\cite{rekimoto2025gazellm} addresses this issue by cropping each frame around the camera wearer's gaze point before it reaches the MLLM. This reduces the MLLM's visual token count by roughly tenfold while maintaining or improving the quality of the video description. The results show that gaze-guided cropping is a viable token-compression strategy. However, it depends entirely on dedicated eye-tracking hardware, which is absent from the camera-equipped smart glasses currently entering the consumer market. 

This dependency raises an obvious question: Could a learned model supply the gaze signal instead, removing the eye tracker from the pipeline entirely? Even an accurate prediction model will disagree with the true fixation point in individual frames, experience lag during rapid saccades, and misjudge attention in visually ambiguous scenes. Whether this imprecision survives the downstream task and whether an MLLM given a predicted-gaze crop describes a scene as well as one given a measured-gaze crop are empirical questions that no prior work has tested directly.

This paper closes that gap. We demonstrate that gaze predicted by a lightweight model can replace hardware-measured gaze in the GazeLLM pipeline. This shows that an eye tracker is not a strict prerequisite for gaze-conditioned egocentric video understanding. Our key contributions are:

\begin{itemize}
\item \textbf{Predicted gaze substitutes for measured gaze in the downstream description task.} The difference between predicted and ground-truth gaze crops is non-significant in all ten cells, spanning two MLLMs and evaluated using three automated metrics and two LLM judges. Equivalence testing confirms this in all ten cells, showing that the two conditions differ by less than half the pooled standard deviation rather than merely failing to detect a difference. In the majority of comparisons, both gaze conditions outperform a center-crop baseline.
\item \textbf{EgoGazeLite, an efficient egocentric gaze predictor for on-device deployment.} At 15.7M parameters and 6.71 GFLOPs, a more than 8-fold FLOP reduction compared to prior egocentric gaze models, EgoGazeLite runs the full gaze-and-crop pipeline end-to-end in real time on consumer accelerator hardware, closing the loop from prediction to on-device deployment.
\end{itemize}

%% file: sections/02_related.tex
\section{Related Work}
\label{sec:related_work}

\paragraph{Egocentric gaze prediction.} There are two broad approaches to predicting gaze in egocentric video. Classical models treat gaze as a function of bottom-up visual saliency ~\cite{itti1998model} and have been extended to egocentric video by incorporating first-person cues, such as camera motion and hand position ~\cite{li2013learning,zhang2017deep}. While these models perform well in free-viewing settings, they struggle when gaze is driven by an unfolding task rather than visual prominence. Huang et al.~\cite{huang2018} addressed this issue by proposing a hybrid CNN-LSTM architecture that combines bottom-up saliency with a top-down attention transition path conditioned on fixation history. We adopt this dual-process structure as our starting point. More recently, Lai et al.~\cite{lai2023} introduced a Transformer-based Global-Local Correlation Module that achieves state-of-the-art accuracy. However, they acknowledge that the computational cost of Transformer architectures may be impractical for on-device AR/VR deployment. Thus, there is a trade-off between accuracy and deployability that EgoGazeLite is designed to navigate. To our knowledge, no prior egocentric gaze predictor has been designed for on-device deployment at the scale we target.

\paragraph{Gaze-guided multimodal LLMs.} Several systems integrate gaze into vision-language models. GazeGPT~\cite{konrad2024gazegpt} uses eye tracking to identify an object in a static scene for an MLLM, and G-VOILA~\cite{wang2024gvoila} combines gaze with voice queries. VQA-MHUG~\cite{sood2021vqamhug} and MULAN~\cite{sood2023mulan} incorporate human gaze into visual question answering. GazeVLM~\cite{chen2025gazevlm} uses measured gaze to crop a region of interest for a VLM on static images, reducing visual tokens by up to 93\% while improving answer quality. The immediate predecessor to this work is GazeLLM~\cite{rekimoto2025gazellm}. It crops egocentric video around measured gaze before passing it to an MLLM, matching or exceeding the quality of full-resolution descriptions at roughly one-tenth the pixel input. We build directly on this approach, replacing measured gaze with EgoGazeLite's predicted gaze while adding equivalence testing to support the substitution claim.

EgoGazeVQA~\cite{peng2025egogazevqa} takes a different route in the same setting, using gaze as a prompt-side signal for longer-form question answering rather than as a cropping mechanism, so the visual token budget is not reduced. Every one of these systems, from static-image VQA to longer-form video, treats measured gaze as a given of the hardware. Whether a predicted gaze signal could substitute for the eye tracker in a cropping pipeline of this kind, closing the loop between prediction and MLLM input on-device, has not been tested. This paper closes that gap for the GazeLLM pipeline.

\paragraph{Reducing tokens after tokenization.} A separate line of work focuses on reducing token count inside the vision encoder itself, rather than reducing the pixel input beforehand. Token merging~\cite{bolya2023tome} progressively combines similar tokens between transformer blocks as a Vision Transformer processes an image, using a lightweight, data-driven matching rule rather than a fixed spatial region. This operates on the token sequence after patch embedding, in contrast to gaze-guided cropping, which reduces the pixel input before it is tokenized. 

%% file: sections/03_method.tex
\section{Method}
\label{sec:method}

\subsection{EgoGazeLite Architecture}
\label{subsec:egogazelite-architecture}
EgoGazeLite builds on Huang et al.'s~\cite{huang2018} dual-process design, identified by Lai et al.~\cite{lai2023} as the strongest non-transformer baseline in their benchmark comparison. A bottom-up path predicts a saliency heatmap from visual input alone; a top-down path predicts an attention heatmap based on recent fixation history; and the two are fused into a single output (Figure~\ref{fig:architecture}). We select this design over Transformer-based approaches because of their impracticality for on-device deployment. Our implementation differs from Huang's in three ways, all of which aim to reduce computational cost: a single shared backbone replaces separate spatial and temporal streams; feature-level temporal difference replaces dense optical flow; and a residual fusion module replaces nonlinear convolutional fusion.

\input{figures/fig_architecture}

\paragraph{Backbone.} The current frame and previous frame are each passed through a single shared EfficientNet-Lite4 \cite{tan2019efficientnet} instance that is initialized from ImageNet-pretrained weights. This replaces Huang's separate spatial and temporal convolutional streams and is the largest single source of parameter reduction relative to their design. The backbone produces a deep feature map of shape $[B, 448, 10, 10]$ per frame, where B is the batch size, 448 is the channel dimension, and $10\times10$ is the spatial resolution. It also produces three intermediate feature maps that are used as skip connections by the saliency decoder below.

\paragraph{Bottom-up path.} A temporal difference module computes
$\Delta F = \text{BatchNorm}(F_t - F_{t-1})$ where $F_t$ and $F_{t-1}$ are the backbone's feature maps for the current and previous frame respectively. This operates at the deep feature resolution of the backbone, replacing Huang's dense optical flow stream with a single subtraction over features that have already been computed. This process requires no additional computational cost beyond the second backbone pass, which is also required by the top-down path. $F_t$ and $\Delta F$ are concatenated channel-wise and passed to a U-Net-style decoder. The decoder upsamples the values to $300\times300$ using skip connections from the three intermediate backbone resolutions. A learnable 2D Gaussian center bias is added to the decoder output before the sigmoid activation, capturing the well-documented center bias of egocentric gaze~\cite{huang2018}. The bottom-up path produces a saliency heatmap $G_t^s \in [B,1,300,300]$ (as logits, since fusion operates in logit space).

\paragraph{Top-down path.} Following Huang et al.~\cite{huang2018}, this path tracks which backbone feature channels were attended to during the previous fixation. It propagates that signature forward during an ongoing fixation and replaces it with a learned LSTM transition when a saccade occurs. A channel weight extractor pools the backbone features at the previous gaze location via RoI-Align. These features are then converted into a per-channel weight vector. A two-layer LSTM predicts the transitioned weights, and an attention weight applier combines the current frame's features with these weights to produce an attention heatmap $G_t^a \in [B,1,300,300]$ (as logits, since fusion operates in logit space). The module's fixation/saccade gate is computed online from recent gaze history via dispersion-threshold identification (I-DT)~\cite{salvucci2000identifying}, rather than a separately trained classifier.

\paragraph{Fusion.} $G_t^s$ and $G_t^a$ are combined by a residual fusion module. The saliency map $G_t^s$ is treated as the base prediction, and a small four-layer convolutional network $R$ produces an additive correction from their concatenation where $\sigma$ denotes the sigmoid function:

\begin{equation}
G_t = \sigma\big(G_t^s + R(\sigma(G_t^s), G_t^a)\big)
\end{equation}

Operating in logit space allows the residual to sharpen or suppress the saliency prediction, depending on whether the attention map agrees. This avoids the saturation that arises from directly combining probability-space heatmaps. This replaces Huang's nonlinear convolutional fusion, which learns the combination end-to-end from a concatenation of the two heatmaps without treating either as a prior. The final gaze coordinate $g_t = \arg\max G_t$ is obtained via argmax over $G_t$.

\paragraph{Training.} EgoGazeLite is trained in three stages that mirror its dual-process structure. In Stage 1, the backbone (with all but the final block frozen at ImageNet weights) and the bottom-up path are trained using distance-weighted binary cross-entropy~\cite{huang2018}. The distance weighting upweights pixels farther from the ground-truth gaze point to counteract the trivial near-zero solution on an otherwise sparse target. Stage 2 trains the top-down path with the bottom-up path frozen. It uses mean squared error between the predicted and ground-truth channel-weight vectors at fixation boundaries. Stage 3 trains the fusion module while jointly fine-tuning the top-down path. It uses the same distance-weighted BCE~\cite{huang2018} objective as Stage 1, but now applies it to the fused heatmap.

\paragraph{Evaluation convention.} We report the average angular error (AAE) using the argmax convention, which is the single point of maximum heatmap value $g_t$. This point is matched with the point prediction that EgoGazeLite outputs during inference. It is also used by the downstream cropping pipeline (Section~\ref{subsec:cropping-pipeline}). This differs from the center-of-mass (CoM) convention, which is used in some parts of the literature. The CoM convention is systematically biased toward the frame center for off-center fixations and therefore understates the average angular error. We report argmax throughout, unless otherwise noted.

\subsection{Gaze-Guided Cropping Pipeline}
\label{subsec:cropping-pipeline}

\paragraph{Dataset.} We use the Ego-Exo4D~\cite{grauman2025egoexo4d} dataset, which is a large-scale egocentric-exocentric video dataset captured with Project Aria Gen 1 glasses~\cite{engel2023aria}. From the dataset, we only use the egocentric RGB stream and its associated gaze annotations. The dataset spans eight task domains: Basketball, Soccer, Bouldering, Dance, and Music (physical activities), and Cooking, Bike Repair, and Health (procedural activities; Health includes CPR training and PCR testing preparation). We train EgoGazeLite on seven of these domains, excluding Basketball, which we held out to probe cross-domain generalization; a systematic evaluation along these lines is left to future work (Section~\ref{sec:conclusion}). The deployed checkpoint is trained on Split A, a subset targeting approximately four hours of video per domain (28 hours total). A second checkpoint trained on the larger Split B (approximately ten hours per domain) yields no meaningful in-distribution accuracy gain over Split A and is not used further.

\paragraph{Crop generation.} We follow Rekimoto's~\cite{rekimoto2025gazellm} cropping protocol. We use each 30\,fps, 1404$\times$1404 source clip and its associated gaze trajectory to produce a 1\,fps, 448$\times$448 cropped video,  sampling every 30th frame and centering a fixed-size window on the gaze coordinate at each sampled timestamp. If the gaze point $g_t$ falls within 224\,px of a frame edge, the crop window is clamped inward to remain within the source frame, preserving its 448$\times$448 size. Three crop conditions are produced per clip: a \emph{predicted} crop, centered on EgoGazeLite's output (Section~\ref{subsec:egogazelite-architecture}) matched to each 1\,fps output tick; a \emph{ground-truth} crop, centered on the Aria eye-tracker's gaze coordinate processed identically; and a \emph{center} crop, fixed at the frame's geometric center regardless of gaze, serving as a spatially-uninformed baseline. Ticks with a ground-truth tracker dropout are omitted rather than interpolated, so ground-truth clips may contain fewer frames than the other two conditions. A fourth, uncropped condition retains the full 1404$\times$1404 frame at the same 1\,fps sampling rate. This condition serves as the reference against which the other three crop conditions are scored. Figure~\ref{fig:crop-example} shows an example frame under all three crop conditions.

\input{figures/fig_crop_example_sushi}

\paragraph{Substitution test set.} We create a held-out evaluation set of 138 clips from Ego-Exo4D. We select clips to match Rekimoto's~\cite{rekimoto2025gazellm} six task categories (Bike Repair, Sushi Preparation, Omelette Preparation, Soccer, PCR Testing Preparation, and CPR Training) with the same number of clips in each category. Since the exact clip identifiers used in Rekimoto's evaluation are unavailable, this is a protocol-faithful reconstruction of equivalent material rather than an exact reproduction. Our set contains one more clip than Rekimoto~\cite{rekimoto2025gazellm} (138 vs.\ 137). We retain all sushi clips. Rekimoto's~\cite{rekimoto2025gazellm} set omits one such clip without explanation. None of the 138 clips appear in EgoGazeLite's training set. The substitution evaluation is therefore held out from the gaze predictor.

\paragraph{Description generation.} Each cropped clip and its full-resolution counterpart are sent to two MLLMs, Gemini~2.5~Flash and Gemini~2.5~Pro~\cite{gemini2025tech}, with a fixed instruction prompt taken verbatim from Rekimoto~\cite{rekimoto2025gazellm} asking the model to produce a written, step-by-step procedure for the depicted activity. The same prompt is used across all four conditions. The generation temperature is set to 0.0 for reproducibility. The resulting descriptions are scored against the full-resolution reference using the metrics and statistical tests reported in Section~\ref{subsec:substitution}.

%% file: figures/fig_architecture.tex
\begin{figure}[b]
\centering
\includegraphics[width=\linewidth]{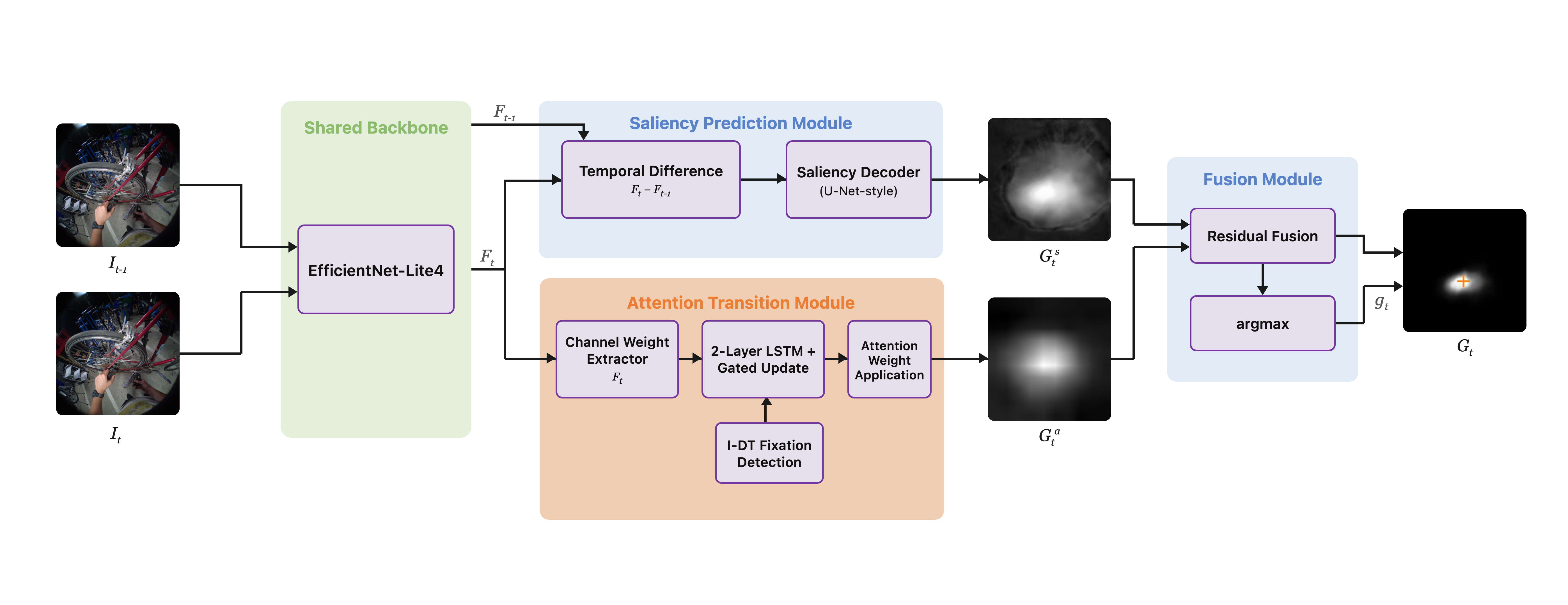}
\caption{EgoGazeLite architecture. Current and previous frames pass through a
shared EfficientNet-Lite4 backbone. The saliency prediction module computes a feature-level temporal difference and decodes it into a bottom-up saliency heatmap $G_t^s$. The attention transition module extracts channel weights at the previous gaze location, propagates them through a gated LSTM (gated by an online I-DT fixation/saccade decision), and applies them to the current frame's features to produce a top-down attention heatmap $G_t^a$. A residual fusion module combines both heatmaps into the final prediction heatmap $G_t$, from which the gaze coordinate ($g_t$) is extracted via argmax.}
\label{fig:architecture}
\end{figure}

%% file: figures/fig_crop_example_sushi.tex
\begin{figure}[t]
\centering
\includegraphics[width=\linewidth]{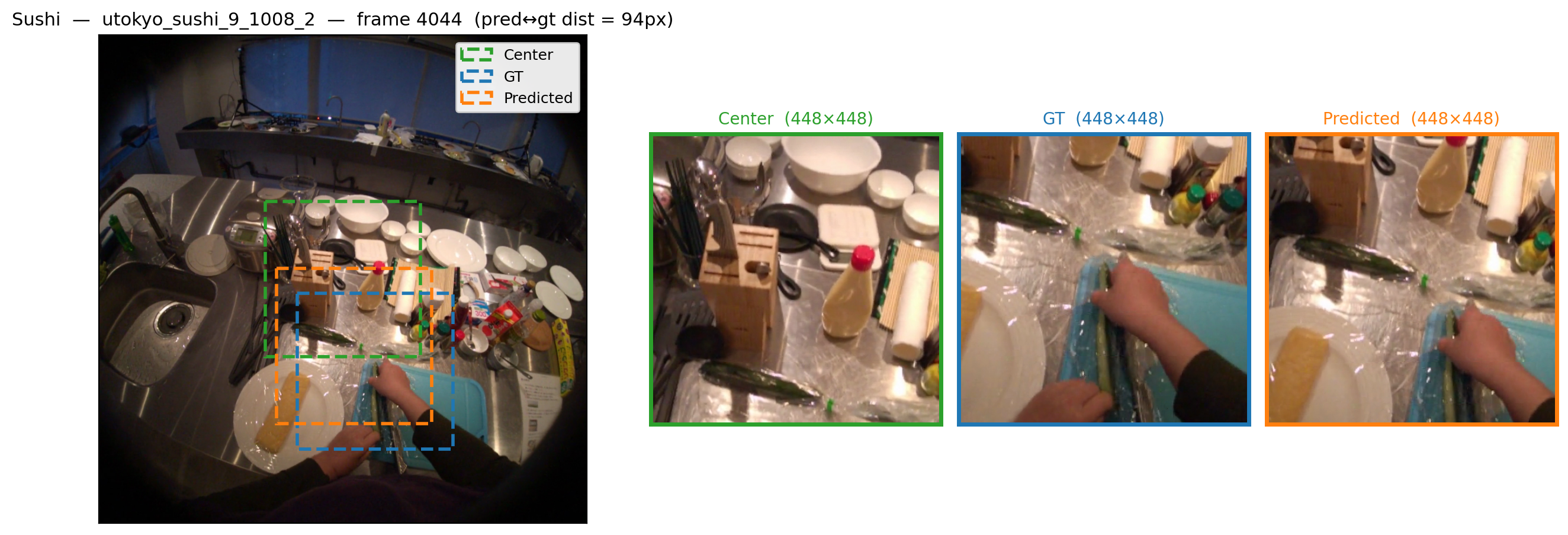}
\caption{Example crop conditions for a single frame (Sushi domain). Center,
ground-truth, and predicted crop windows are shown on the full frame (left) and
as extracted 448$\times$448 crops (right). In this frame, the predicted and
ground-truth crops overlap substantially (pred$\leftrightarrow$GT distance:
94\,px).}
\label{fig:crop-example}
\end{figure}

%% file: sections/04_experiments.tex
\section{Experiments}
\label{sec:experiments}

This section presents two experiments that, together, demonstrate the two contributions of the paper. Section~\ref{subsec:egogazelite-eval} evaluates the accuracy, efficiency, and on-device latency of EgoGazeLite, demonstrating that the gaze-and-crop pipeline built around EgoGazeLite runs in real time on edge hardware (C2). Section~\ref{subsec:substitution} tests whether the gaze produced by this predictor can substitute for hardware-measured gaze in a downstream multimodal-LLM video-description task. It compares description quality across crop conditions and tests for both difference and equivalence to determine substitutability (C1).

\subsection{EgoGazeLite: Accuracy, Efficiency, and On-Device Latency}
\label{subsec:egogazelite-eval}

\paragraph{Accuracy.} We evaluate the performance of the deployed checkpoint, which was trained using the Ego-Exo4D Split A dataset in Section~\ref{subsec:egogazelite-architecture}, on its in-distribution validation set across seven training domains: Cooking, Soccer, Health, Bike Repair, Dance, Bouldering, and Music. We report the Area Under the Curve (AUC), computed as a single-threshold saliency AUC following the implementation of Huang et al.~\cite{huang2018}, and Average Angular Error (AAE)~\cite{riche2013saliency}. Figure~\ref{fig:heatmap-domains} shows the predicted and ground-truth gaze locations for four of these domains. Aggregated across all seven domains, EgoGazeLite achieves an AUC of 0.9655, an AAE (argmax convention, Section~\ref{subsec:egogazelite-architecture}) of 8.33°, and a pixel distance of 26.35 px. The per-domain AUC ranges from 0.9545 (Music) to 0.9705 (Cooking), with a spread of 0.016. The three procedural domains cluster within 0.0037 AUC. The physical domains are more variable, ranging from 0.9702 (Soccer) to 0.9545 (Music), with Soccer performing comparably to the procedural domains despite its task type. The full per-domain results are in Table~\ref{tab:egoexo4d}. We report these numbers to characterize the deployed checkpoint. The relevant validation of accuracy is the downstream substitution test in Section~\ref{subsec:substitution}, which measures whether this level of gaze prediction preserves MLLM description quality. This is a task-grounded criterion that a raw metric comparison against prior egocentric gaze models (Huang et al.~\cite{huang2018}, Lai et al.~\cite{lai2023}) would not address.

\input{tables/tab_egoexo4d}

\input{figures/fig_per_domain}

\paragraph{Efficiency.} Table~\ref{tab:capacity} compares parameter count and FLOPs against the two prior egocentric gaze predictors that motivate this work. EgoGazeLite uses 15.7M parameters, which is 3.2 times fewer than Huang et al.'s dual-stream CNN-LSTM~\cite{huang2018} and 4.5 times fewer than Lai et al.'s Global-Local Correlation Transformer~\cite{lai2023}. The FLOPs gap is wider: 6.71\,GFLOPs, which is an 8.6-fold reduction relative to Huang and a 14-fold reduction relative to Lai. Both Huang's optical-flow stream and Lai's clip-level attention over eight-frame windows at 256$\times$256 resolution inflate the computational cost independently of the parameter count. EgoGazeLite instead processes each frame through a single shared backbone (Section~\ref{subsec:egogazelite-architecture}), avoiding both sources of overhead.

\input{tables/tab_capacity}

\paragraph{On-device latency.} We measure the full gaze-and-crop pipeline (Section~\ref{subsec:cropping-pipeline}) end-to-end on an iPhone 15 Pro. This is the class of device that would host the MLLM connection in a realistic smart-glasses deployment. Therefore, the crop step naturally sits on the phone between the glasses' video stream and the MLLM API. Following 10 warm-up runs on synthetic 4K input, the pipeline completes in an average of 21.6 ms per frame (approximately 46 FPS) over 100 iterations, with a P95 of 30.8 ms, excluding the downstream MLLM call. The EgoGazeLite forward pass on the Neural Engine accounts for 8.9 ms (approximately 41\%). The remaining time is divided between the initial cropping and downscaling (8.3 ms), extracting the cropped image at its native resolution (3.9 ms), and normalization (0.4 ms). The model is not the bottleneck; the surrounding image-processing steps are. For the evaluation in Section~\ref{subsec:substitution}, the MLLM input is sampled at one FPS, meaning the crop step runs more than 40 times faster than the downstream sampling rate requires, and the pipeline can also support sampling rates of up to 30 FPS. This puts the pipeline well within the frame-rate needs of the downstream MLLM. The reduction in visual tokens by about 10 times propagates directly to the phone-to-MLLM upload, reducing bandwidth for every downstream call, for a small local latency cost.

\subsection{Predicted Gaze as a Substitute for Measured Gaze}
\label{subsec:substitution}
We test whether EgoGazeLite's predictions can replace real eye-tracker gaze in the cropping pipeline. Each clip is processed four ways: full frame, ground-truth gaze crop, predicted gaze crop, and center crop. Descriptions from the three $448\times448$ crops, generated by Gemini 2.5 Flash and Gemini 2.5 Pro~\cite{gemini2025tech}, are scored against the full-frame description using three automated metrics and two LLM judges. BLEU~\cite{papineni2002bleu} measures n-gram precision overlap with the reference description, ROUGE-L~\cite{lin2004rouge} measures longest common subsequence overlap, and SBERT~\cite{reimers2019sbert} measures cosine similarity between sentence embeddings; all three are naturally bounded on a 0–1 scale. The two LLM judges, Claude Sonnet 4.6~\cite{anthropic2026sonnet} and GPT-4o~\cite{openai2024gpt4o}, instead rate each description from 0 to 100 on how completely it covers the key information of the reference description, following Rekimoto's~\cite{rekimoto2025gazellm} judge prompt. Six of the 828 cropped-condition runs failed to generate; all of those failed runs were on Flash. Those clips are dropped listwise, leaving 134–138 per cell.

\paragraph{Description quality by condition.} Table~\ref{tab:quality} reports the mean quality for each condition, MLLM, and metric. The predicted and ground-truth crops outperform the center crop on every metric and both MLLMs. They also track each other closely. On Pro, the difference between the predicted and ground-truth means is at most 0.002 for BLEU and ROUGE-L, and 0.006 for SBERT. On Flash, the difference is under 0.014 for BLEU and 0.016 for ROUGE-L. In four cells, the predicted mean is higher: Pro ROUGE-L (0.459 vs. 0.457), Pro Claude (71.42 vs. 68.15), Flash GPT-4o (75.52 vs. 75.19) and Pro GPT-4o (80.54 vs. 79.42). Pro scores higher than Flash everywhere, and the pattern of predicted versus ground truth is the same on both.

\input{tables/tab_quality}

\paragraph{Difference tests.} We run paired t-tests over three contrasts for each cell: predicted vs. ground truth, predicted vs. center, and ground truth vs. center. Holm correction~\cite{holm1979} is applied to each cell's family of three (Table~\ref{tab:tests}). The predicted versus ground truth contrast is non-significant in all ten cells. The smallest corrected p-value is 0.054 (Flash ROUGE-L), and the absolute value of the standardized effect sizes ($dz$) stays below 0.17 everywhere, well within Cohen's convention for a small effect ($dz < 0.2$), including the three cells where the predicted value is higher. Figure~\ref{fig:per-task-judge} shows the LLM-judge scores per task, where the pattern holds across categories: predicted and ground-truth crops track each other closely on every task, with soccer showing the smallest advantage of gaze conditions over center. The automated metrics (BLEU, ROUGE-L, SBERT) show the same aggregate pattern (Table~\ref{tab:quality}) but are not broken out per task. The gaze-versus-center contrasts reach $p_{\text{Holm}}$ <.001 in 16 of the 20 cells, with $d_z$ ranging from 0.19 to 0.68. Only the predicted-versus-center contrast for Flash BLEU misses correction ($p_{\text{Holm}}$ = 0.054); its ground-truth counterpart clears it.

\input{figures/fig_per_task_judge}

\paragraph{Equivalence tests.}  Since a non-significant difference does not imply equivalence, we conduct two one-sided tests (TOST~\cite{lakens2017tost}) on the predicted versus ground truth contrast. No domain-specific equivalence margin exists for this task. Following Lakens' recommendation to use standardized effect-size conventions as the equivalence margin when no field-established benchmark is available~\cite{lakens2017tost}, we set the primary margin at Cohen's medium-effect threshold~\cite{cohen1988}, $0.5\times$ the pooled standard deviation of the ground-truth and predicted score distributions per metric, rather than the more permissive large-effect threshold ($0.8\times$ pooled SD). We additionally report equivalence under Cohen's stricter small-effect threshold ($0.2\times$ pooled SD) as a sensitivity check. As a validity check, we apply the same margin to the gaze versus center contrast, a comparison with a known substantial effect. Equivalence is correctly rejected in all ten cells, confirming that the equivalence test is sensitive enough to detect a real difference when one exists. Holm correction is applied across the ten tests. Equivalence holds in all ten cells (see Table~\ref{tab:tests}). Tightening the margin to 0.2 times the pooled standard deviation yields equivalence in four of the ten cells: Flash GPT-4o, Pro BLEU, Pro ROUGE-L, and Pro GPT-4o. Figure~\ref{fig:tost-forest} shows the standardized effect sizes and confidence intervals for all ten cells.

\input{tables/tab_tests}

\input{figures/fig_tost_forest}

\paragraph{Robustness to judge choice.} GPT-4o scores about ten points higher than Claude in every category (80.54 vs. 71.42 for Pro Predicted, for example), yet the two models agree on the order. The clip-level Spearman correlation is +0.81 overall, computed separately for each MLLM–condition subset. It ranges from +0.72 (Flash, ground-truth and predicted) to +0.89 (Pro center). This correlation is significant at $p < 10^{-23}$. Swapping the judge changes the absolute scores but not the ranking of the three crops.

%% file: tables/tab_egoexo4d.tex
\begin{table}[t]
\centering
\caption{In-distribution gaze prediction accuracy of the deployed EgoGazeLite
checkpoint (trained on Ego-Exo4D Split~A, Section~\ref{sec:method}) across seven training domains. Average Angular Error (AAE) uses the argmax convention (Section~\ref{sec:method}).}
\label{tab:egoexo4d}
\begin{tabular}{lccc}
\toprule
Domain & AUC & AAE (\textdegree) & Pixel distance \\
\midrule
Cooking        & 0.9705 & 7.58  & 24.46 \\
Soccer         & 0.9702 & 7.67  & 23.68 \\
Health         & 0.9684 & 7.73  & 25.51 \\
Bike Repair    & 0.9668 & 8.03  & 25.76 \\
Dance          & 0.9587 & 9.34  & 28.97 \\
Bouldering  & 0.9566 & 9.64  & 30.16 \\
Music          & 0.9545 & 10.23 & 30.76 \\
\midrule
\textbf{Overall} & \textbf{0.9655} & \textbf{8.33} & \textbf{26.35} \\
\bottomrule
\end{tabular}
\end{table}

%% file: figures/fig_per_domain.tex
\begin{figure}[t]
\centering
\includegraphics[width=\linewidth]{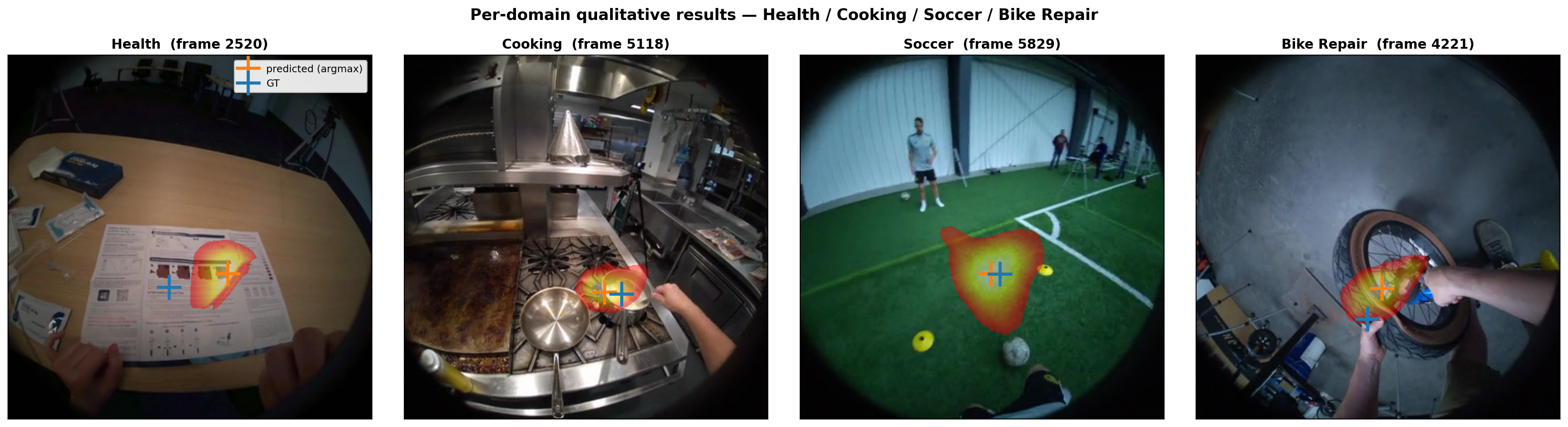}
\caption{Predicted (orange) versus ground-truth (blue) gaze locations across four
Ego-Exo4D domains, overlaid on EgoGazeLite's predicted heatmap. Predicted and
ground-truth points align closely in Cooking, Soccer, and Bike Repair; Health
shows a larger deviation, illustrating a case where prediction is less precise.}
\label{fig:heatmap-domains}
\end{figure}

%% file: tables/tab_capacity.tex
\begin{table}[t]
\centering
\caption{Model capacity comparison. EgoGazeLite uses substantially fewer parameters and FLOPs than prior egocentric gaze predictors.}
\label{tab:capacity}
\begin{tabular}{lcc}
\toprule
Model & Parameters & FLOPs \\
\midrule
EgoGazeLite            & 15.7M & 6.71\,G  \\
Huang et al.~\cite{huang2018} & 51.0M & 57.76\,G \\
Lai et al.~\cite{lai2023}     & 70.2M & 94.22\,G \\
\bottomrule
\end{tabular}
\end{table}

%% file: tables/tab_quality.tex
\begin{table}[t]
\centering
\caption{Description quality by crop condition, MLLM, and metric (mean\,$\pm$\,SE).
Automated metrics are on a 0--1 scale; LLM judges on a 0--100 scale. Bold indicates
the best-performing condition per column.}
\label{tab:quality}
\resizebox{\linewidth}{!}{%
\setlength{\tabcolsep}{4pt}
\footnotesize
\begin{tabular}{llccccc}
\toprule
MLLM & Condition & BLEU & ROUGE-L & SBERT & Claude & GPT-4o \\
\midrule
\multirow{3}{*}{Gemini 2.5 Flash}
 & Center       & $0.213{\scriptstyle\pm0.009}$ & $0.334{\scriptstyle\pm0.008}$ & $0.868{\scriptstyle\pm0.010}$ & $54.25{\scriptstyle\pm2.24}$ & $65.63{\scriptstyle\pm1.96}$ \\
 & Ground-truth & $\mathbf{0.248}{\scriptstyle\pm0.010}$ & $\mathbf{0.376}{\scriptstyle\pm0.009}$ & $\mathbf{0.912}{\scriptstyle\pm0.005}$ & $\mathbf{68.04}{\scriptstyle\pm2.14}$ & $75.19{\scriptstyle\pm1.71}$ \\
 & Predicted    & $0.234{\scriptstyle\pm0.009}$ & $0.361{\scriptstyle\pm0.009}$ & $0.904{\scriptstyle\pm0.005}$ & $64.28{\scriptstyle\pm2.16}$ & $\mathbf{75.52}{\scriptstyle\pm1.67}$ \\
\midrule
\multirow{3}{*}{Gemini 2.5 Pro}
 & Center       & $0.244{\scriptstyle\pm0.007}$ & $0.398{\scriptstyle\pm0.007}$ & $0.876{\scriptstyle\pm0.010}$ & $53.79{\scriptstyle\pm2.24}$ & $68.04{\scriptstyle\pm2.28}$ \\
 & Ground-truth & $\mathbf{0.300}{\scriptstyle\pm0.008}$ & $0.457{\scriptstyle\pm0.008}$ & $\mathbf{0.911}{\scriptstyle\pm0.007}$ & $68.15{\scriptstyle\pm2.32}$ & $79.42{\scriptstyle\pm1.94}$ \\
 & Predicted    & $\mathbf{0.300}{\scriptstyle\pm0.008}$ & $\mathbf{0.459}{\scriptstyle\pm0.008}$ & $0.905{\scriptstyle\pm0.007}$ & $\mathbf{71.42}{\scriptstyle\pm2.12}$ & $\mathbf{80.54}{\scriptstyle\pm1.89}$ \\
\bottomrule
\end{tabular}%
}
\end{table}

%% file: figures/fig_per_task_judge.tex
\begin{figure}[t]
\centering
\includegraphics[width=0.85\linewidth]{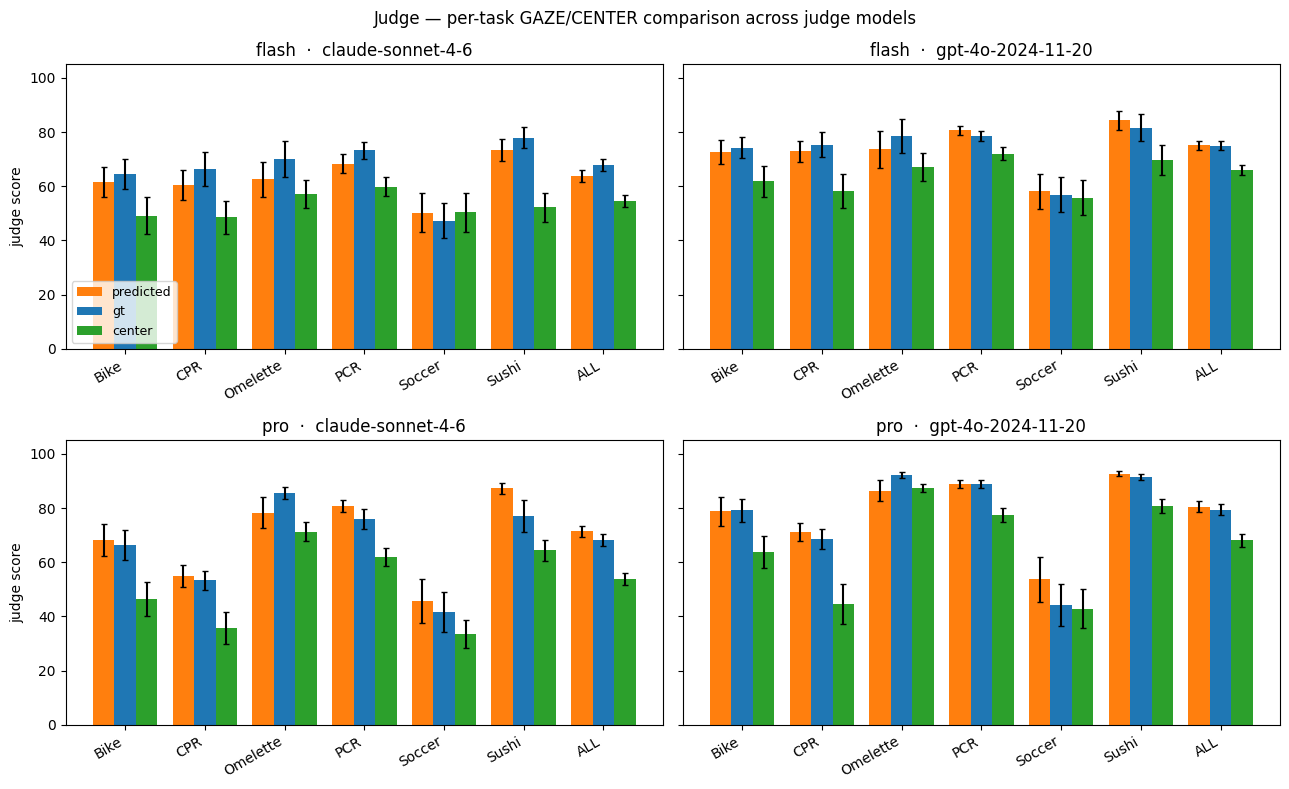}
\caption{Per-task description quality by crop condition, across the two MLLMs (rows: Gemini 2.5 Flash, Gemini 2.5 Pro) and two LLM judges (columns: Claude Sonnet 4.6, GPT-4o). Predicted-gaze (orange) and ground-truth-gaze (blue) crops track each other on every task, while center crops (green) score consistently lower. The rightmost 'ALL' bar in each panel is the mean across all tasks. Error bars are $\pm 1$ SE. Soccer shows the smallest gap between gaze-conditioned and center crops.}
\label{fig:per-task-judge}
\end{figure}

%% file: tables/tab_tests.tex
\begin{table}[t]
\centering
\caption{Predicted vs.\ ground-truth gaze: difference and equivalence tests. Difference columns report the mean paired difference, Cohen's $d_z$ (mean paired difference divided by SD of paired differences), and the Holm-corrected paired-$t$-test $p$-value. The TOST $p_{\mathrm{Holm}}$ column reports the Holm-corrected equivalence test result; standardized effect sizes and 90\% confidence intervals for these tests are shown in Figure~\ref{fig:tost-forest}. Margins are set at $0.5\times$ (primary) and $0.2\times$ (sensitivity) the pooled SD of the ground-truth and predicted score distributions per metric, following Cohen's medium/small effect-size convention~\cite{cohen1988}.}
\label{tab:tests}
\resizebox{\linewidth}{!}{%
\setlength{\tabcolsep}{4pt}
\footnotesize
\begin{tabular}{llcccccc}
\toprule
 & & \multicolumn{3}{c}{Difference} & \multicolumn{3}{c}{Equivalence (TOST)} \\
\cmidrule(lr){3-5}\cmidrule(lr){6-8}
MLLM & Metric & Mean diff & $d_z$ & $p_{\mathrm{Holm}}$ & TOST $p_{\mathrm{Holm}}$ & Primary & Sensitivity \\
\midrule
\multirow{5}{*}{Flash}
 & BLEU    & $-0.014$ & $-0.12$ & $0.163$ & $<0.001$ & \checkmark & $\times$ \\
 & ROUGE-L & $-0.016$ & $-0.17$ & $0.054$ & $<0.001$ & \checkmark & $\times$ \\
 & SBERT   & $-0.008$ & $-0.15$ & $0.085$ & $<0.001$ & \checkmark & $\times$ \\
 & Claude  & $-3.76$  & $-0.15$ & $0.076$ & $<0.001$ & \checkmark & $\times$ \\
 & GPT-4o  & $+0.34$  & $+0.02$ & $0.795$ & $<0.001$ & \checkmark & \checkmark \\
\midrule
\multirow{5}{*}{Pro}
 & BLEU    & $-0.000$ & $-0.01$ & $0.950$ & $<0.001$ & \checkmark & \checkmark \\
 & ROUGE-L & $+0.002$ & $+0.02$ & $0.804$ & $<0.001$ & \checkmark & \checkmark \\
 & SBERT   & $-0.006$ & $-0.10$ & $0.223$ & $<0.001$ & \checkmark & $\times$ \\
 & Claude  & $+3.27$  & $+0.15$ & $0.083$ & $<0.001$ & \checkmark & $\times$ \\
 & GPT-4o  & $+1.12$  & $+0.08$ & $0.352$ & $<0.001$ & \checkmark & \checkmark \\
\bottomrule
\end{tabular}%
}
\end{table}

%% file: figures/fig_tost_forest.tex
\begin{figure}[t]
\centering
\includegraphics[width=0.85\linewidth]{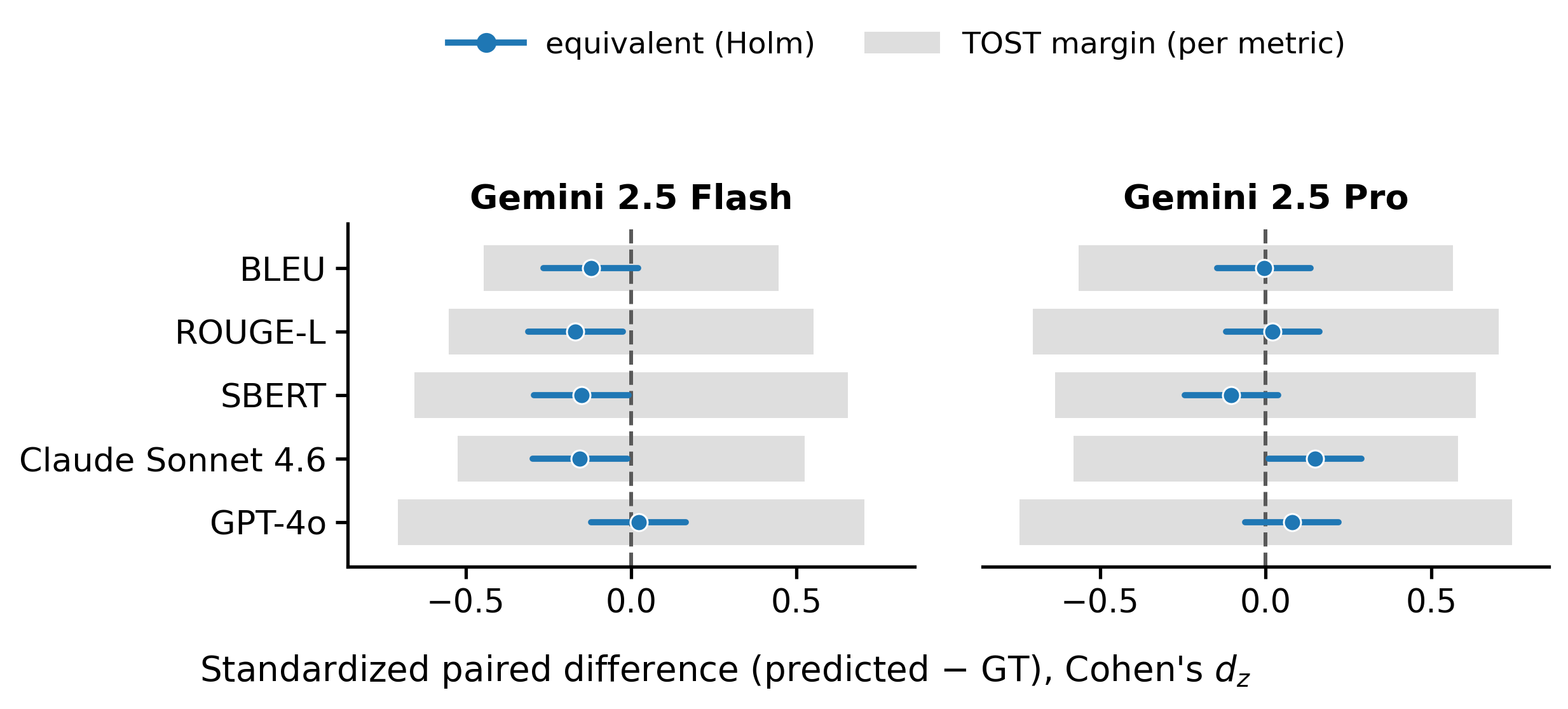}
\caption{Equivalence testing (TOST) of predicted-vs-ground-truth gaze crops, Holm-corrected across all ten cells. Points show the standardized mean paired difference (Cohen's $d_z$), bars show 90\% confidence intervals. The shaded band is an approximate $d_z$-scale mapping of the TOST margins ($0.5\times$ the pooled SD of the ground-truth and predicted score distributions per metric, Table~\ref{tab:tests}).}
\label{fig:tost-forest}
\end{figure}

%% file: sections/05_conclusion.tex
\section{Conclusion}
\label{sec:conclusion}

We demonstrated that gaze predicted by a lightweight model can substitute for hardware-measured gaze in a gaze-guided MLLM video-cropping pipeline, closing a gap left by prior work that relied entirely on dedicated eye-tracking hardware. Across two MLLMs, three automated metrics, and two LLM judges, predicted and ground-truth gaze crops showed no significant difference in downstream description quality. Equivalence was confirmed in ten out of ten cases, and both significantly outperformed a center-crop baseline. The substitution of predicted for measured gaze is made practical by EgoGazeLite, a lightweight egocentric gaze predictor that runs the full gaze-and-crop pipeline end-to-end in real time on consumer accelerator hardware, at a fraction of the parameters and FLOPs of prior egocentric gaze prediction models. Together, these results remove the need for eye-tracking hardware for token-efficient, gaze-conditioned egocentric video understanding with MLLMs.

Several limitations bound these results. A direct AAE/AUC comparison to Huang et al.~\cite{huang2018} and Lai et al.~\cite{lai2023} under a matched training split and evaluation protocol was not performed; the three models were trained on different data, and a controlled head-to-head comparison is left to future work. EgoGazeLite is trained and evaluated in-distribution across seven Ego-Exo4D domains, and whether the predictor or the downstream substitution hold up under domain shift to unseen activities is untested. The three main architectural changes described in Section~\ref{subsec:egogazelite-architecture} (shared backbone, feature-difference temporal signal, residual fusion) are not individually ablated; their combined effect is validated against the downstream substitution task. The substitution evaluation uses two MLLMs from a single model family (Gemini 2.5 Flash and Pro); whether the finding generalizes to other model families, training data, and architectures is unknown. Description quality is scored with automated metrics and LLM judges rather than human raters, where prior work in this area has included both. We measure the phone-side pipeline, on an iPhone 15 Pro, and the resulting reduction in pixel area from cropping. We do not directly measure the three deployment-relevant quantities this reduction is intended to improve: end-to-end application latency, upload bandwidth, and MLLM-side visual token count. Each depends on factors outside the pipeline studied here: network conditions and MLLM provider capacity for latency, video encoding for bandwidth, and the provider's tokenization scheme for token count. Smart-glasses-side deployment likewise requires additional validation on target hardware.

Closing these gaps is natural future work. A controlled accuracy comparison against Huang et al.~\cite{huang2018} and Lai et al.~\cite{lai2023} under an identical training split would clarify how much prediction accuracy is traded for EgoGazeLite's efficiency gains. A systematic leave-one-domain-out evaluation across all eight Ego-Exo4D domains would establish how consistently gaze-prediction accuracy and downstream substitution quality transfer under domain shift. An ablation isolating each of the three architectural changes would clarify their individual contributions to the accuracy-efficiency trade-off. A human-rater study would provide an evaluation independent of the automated metrics and LLM judges used here. Validation on smart-glasses hardware would confirm the latency result under the tighter thermal and power constraints of the on-glasses compute path. Measuring end-to-end application latency, upload bandwidth, and MLLM-side visual token count directly, under realistic network and provider conditions, would establish how much of the phone-side efficiency and pixel-area reduction demonstrated here translates into deployment-relevant savings in practice. A direct accuracy-compute comparison against post-tokenization methods such as token merging~\cite{bolya2023tome}, and combining the two approaches, would clarify how gaze-guided cropping complements other routes to MLLM efficiency.

If the substitution result holds beyond the datasets, model families, and task domains tested here, it points to a broader shift: gaze-conditioned egocentric video understanding no longer requires dedicated eye-tracking hardware and becomes available to any first-person camera, not just tracker-equipped wearables.